\documentclass[runningheads]{llncs}

\usepackage[table,xcdraw]{xcolor}
\usepackage[english]{babel}
\usepackage[utf8x]{inputenc}
\usepackage{amsmath}
\usepackage{todonotes}
\usepackage[hyphens]{url}
\usepackage{hyperref}
\usepackage{enumitem}
\usepackage{booktabs}
\usepackage{multirow}
\usepackage{makecell}
\usepackage[table,xcdraw]{xcolor}
\usepackage{cite}
\usepackage[flushleft]{threeparttable}
\usepackage{pgfplots}
\usepackage[export]{adjustbox} 
\usepackage{subfig}
\pgfplotsset{compat=1.5} 

\newcommand\schema[1]{{\normalfont\fontfamily{cmvtt}\selectfont #1}}

\begin{document}

\title{EventKG: A Multilingual Event-Centric Temporal Knowledge Graph}

\author{Simon Gottschalk \and Elena Demidova}

\institute{L3S Research Center, Leibniz Universit\"at Hannover, Hannover, Germany \\
 \email{\{gottschalk,demidova\}@L3S.de}
 } 

\maketitle    

\begin{abstract}
One of the key requirements to facilitate semantic analytics of  
information regarding contemporary and historical events 
on the Web, in the news and in social media is the availability of reference knowledge repositories containing comprehensive representations of events and temporal relations.
Existing knowledge graphs, with popular examples including DBpedia, YAGO and Wikidata, focus mostly on entity-centric information and are insufficient in terms of their coverage and completeness with respect to events and temporal relations.
EventKG presented in this paper is a multilingual event-centric temporal knowledge graph that addresses this gap. 
EventKG incorporates over 690 thousand contemporary and historical events and over 2.3 million temporal relations extracted from several large-scale knowledge graphs and semi-structured sources and makes them available through a canonical representation.

\end{abstract}


\noindent
\textbf{Resource type:} Dataset\\
\textbf{Permanent URL:} \url{http://eventkg.l3s.uni-hannover.de} \\


\section{Introduction}
\label{sec:intro}

\textit{Motivation:}
The amount of event-centric information regarding contemporary and historical events of global importance, such as Brexit, the 2018 Winter Olympics and the Syrian Civil War, constantly grows on the Web, in the news sources and within social media. 
Efficiently accessing and analyzing large-scale event-centric and temporal information is crucial for a variety of real-world applications in the fields of Semantic Web, NLP and Digital Humanities. 
In Semantic Web and NLP, these applications include Question Answering \cite{HoffnerWMULN17} and timeline generation \cite{Althoff:2015}.
In Digital Humanities, multilingual event repositories can facilitate cross-cultural studies that aim to analyze language-specific and community-specific views on historical and contemporary events (examples of such studies can be seen in \cite{GottschalkDBR17}, \cite{Rogers:2013}). 
Furthermore, event-centric knowledge graphs can facilitate the reconstruction of histories as well as networks of people and organizations over time \cite{ROSPOCHER2016132}.
One of the pivotal pre-requisites to facilitate effective analytics of contemporary and historical events is the availability of knowledge repositories providing reference information regarding events, involved entities and their temporal relations (i.e. relations valid over a time period). 

\textit{Limitations of the existing sources of event-centric and temporal information:}
Currently, event representations and temporal relations are spread across heterogeneous sources. 
First, large-scale knowledge graphs (KGs) (i.e. graph-based knowledge repositories \cite{Faerber:2016} such as Wikidata \cite{Erxleben:2014}, DBpedia \cite{dbpedia-swj}, and YAGO \cite{Mahdisoltani:2014}) typically focus on entity-centric knowledge. Event-centric information included in these sources is often not clearly identified as such, can be incomplete and is mostly restricted to named events and encyclopedic knowledge.
For example, as it will be discussed later in Section \ref{sec:characteristics}, of $322,669$ events included in EventKG, only $18.70\%$ are classified using the \schema{dbo:Event} class in the English DBpedia. Furthermore, event descriptions in the existing knowledge graphs often lack the key properties such as times and locations. For example, only $33\%$ of the events in Wikidata provide temporal and $11.70\%$ spatial information.
Second, a variety of manually curated semi-structured sources (e.g. Wikipedia Current Events Portal (WCEP) \cite{Tran:2014} and multilingual Wikipedia event lists) contain information on contemporary events. However, the lack of structured representations of events and temporal relations in these sources hinders their direct use in real-world applications through semantic technologies. 
Third, recently proposed knowledge graphs containing contemporary events extracted from unstructured news sources (such as \cite{ROSPOCHER2016132}) are potentially highly noisy (e.g. \cite{ROSPOCHER2016132} reports an extraction accuracy of $0.55$) 
and are not yet widely adopted.
Finally, the sources of event-centric information that can potentially be explored in future work are Web markup \cite{tempelmeier2018} and event-centric Web crawls \cite{GossenDR15}.
Overall, a comprehensive integrated view on contemporary and historical events and their temporal relations usable for real-world applications is still missing. 
The provision of EventKG will help to overcome these limitations.

\textit{EventKG \& advances to the state of the art:}
EventKG presented in this paper takes an important step to facilitate a global view on events and temporal relations currently spread across entity-centric knowledge graphs and manually curated
semi-structured sources. 
EventKG extracts and integrates this knowledge in an efficient light-weight fashion, enriches 
it with additional features, such as indications of relation strengths and event popularity, adds provenance information 
and makes all this information available through a canonical representation. 
EventKG follows best practices in data publishing and reuses existing data models and vocabularies (such as Simple Event Model \cite{VanHage:2011} and the DBpedia ontology) to facilitate its efficient reuse in real-world applications through the application of semantic technologies and open standards (i.e. RDF and SPARQL).  
EventKG currently includes data sources in five languages -- English (en), German (de), French (fr), Russian (ru) and Portuguese (pt) -- and is extensible. 
The main contributions of EventKG are as follows: 
\begin{itemize}
\item A multilingual RDF knowledge graph incorporating over $690$ thousand events and over $2.3$ million temporal relations in V1.1 extracted from 
several large-scale entity-centric knowledge graphs (i.e. Wikidata, DBpedia in five language editions and YAGO), as well as WCEP and Wikipedia event lists in five languages. In the following, we refer to these sources used to populate EventKG as \textit{the reference sources}. 
The key features of EventKG include:
\begin{itemize}
\item provision of event-centric information (including historical and contemporary events) and temporal relations using a canonical representation; 
\item light-weight integration and fusion of event representations and relations originating from heterogeneous reference sources; 
\item higher coverage and completeness of event representations compared to the individual reference sources (see Section \ref{sec:characteristics}); 
\item provision of interlinking information, to facilitate e.g. an assessment of relation strength and event popularity; 
\item provenance for all information contained in EventKG. 
\end{itemize}
\item An open source extraction framework to extract and maintain up-to-date versions of EventKG, extensible to further languages and reference sources. 
\end{itemize}

\textit{Comparison to other existing resources:}
To the best of our knowledge, currently there are no dedicated knowledge graphs aggregating event-centric information and temporal relations for historical and contemporary events directly comparable to EventKG. 
The heterogeneity of data models and vocabularies for event-centric and temporal information (e.g. \cite{Shaw:2013,ROSPOCHER2016132,VanHage:2011,Guha:2011}), the large scale of the existing knowledge graphs, in which events play only an insignificant role, and the lack of clear identification of event-centric information, makes it particularly challenging to identify, extract, fuse and efficiently analyze event-centric and temporal information and make it accessible to real-world applications in an intuitive and unified way.
Through the light-weight integration and fusion of event-centric and temporal information from different sources, EventKG enables to increase coverage and completeness of this information. For example, EventKG increases the coverage of locations and dates for Wikidata events it contains by $14.43\%$ and $17.82\%$, correspondingly (see Table \ref{tab:named_events_comparison} in Section \ref{sec:characteristics} for more detail).
Furthermore, existing sources lack structured information to judge event popularity and relation strength as provided by EventKG -- the characteristic that gains the key relevance given the rapidly increasing amount of event-centric and temporal data on the Web and the information overload. 

\section{Relevance}
\label{sec:relevance}

\textit{Relevance to the Semantic Web community and society:}
Our society faces an unprecedented number of events that impact multiple communities across language and community borders. In this context, efficient access to, as well as effective disambiguation of, and analytics of event-centric multilingual information originating from different sources, as facilitated by EventKG, is of utmost importance for several scientific communities, including Semantic Web, NLP and Digital Humanities. 
In the context of the Semantic Web community, application areas of EventKG include event-centric Question Answering and ranking-based timeline generation that requires assessment of event popularity and relation strength.
In Digital Humanities, EventKG as a multilingual event-centric repository can provide a unique source for cross-cultural and cross-lingual event-centric analytics (e.g. illustrated in \cite{GottschalkDBR17}, \cite{Rogers:2013}), while reducing barriers of data extraction, integration and fusion.   

\textit{Relevance for Question Answering applications:} 
In the field of Question Answering (QA) \cite{HoffnerWMULN17}, the current focus of research is on the generation of formal query expressions (e.g. in the SPARQL query language) from user queries posed in a natural language as well as interactive approaches for QA and semantic search \cite{Zheng:2017}, \cite{Demidova:2013QC}. Currently, research is mostly performed on questions that can be answered using popular entity-centric knowledge graphs such as DBpedia. With the provision of EventKG, it will become possible to train QA approaches for event-related questions, e.g. \textit{``Which events related to Bill Clinton happened in Washington in 1980?''} and ranking-based questions, e.g. \textit{``What are the most important events related to Syrian Civil War that took place in Aleppo?''}

\textit{Relevance for timeline generation applications:}
Timeline generation is an active research area \cite{Althoff:2015}, where the focus is to generate a timeline (i.e. a chronologically ordered selection) of events and temporal relations for entities from a knowledge graph.
EventKG can facilitate the generation of detailed timelines containing complementary information originating from different sources, potentially resulting in more complete timelines and event representations. For example, Table \ref{tab:wwii_timeline}
illustrates an excerpt from the timeline for the query \textit{``What were the sub-events of the World War II between February 12 and February 28, 1941?''} generated using EventKG. 
The first event in the timeline in Table \ref{tab:wwii_timeline} (``Erwin Rommel arrives in Tripoli'') extracted from an English Wikipedia event list (``1941 in Germany'') is not contained in any of the reference knowledge graphs used to populate EventKG (Wikidata, DBpedia, and YAGO). 
The reference sources of the other three events include complementary information. For example, while the ``Action of 27 February 1941'' is assigned a start date in Wikidata, it is not connected to the World War II in that source.

\begin{table}[!t]
      \centering
      \begin{threeparttable}
    \caption{All sub-events of the World War II in EventKG that started between February 12 and February 28, 1941.}
          \label{tab:wwii_timeline}

      \footnotesize
      \centering
\begin{tabular}{|l|l|p{5.5cm}|}
\hline
\textbf{\makecell{Start Date}} & \textbf{Sources} & \textbf{Description}                                                                                                                                                                         \\ \hline
Feb 12 & Wikipedia event lists & Erwin Rommel arrives in Tripoli.                                                                                                                                        \\ \hline
Feb 17 & YAGO, DBpedia & Battle of Trebeshina. \\ \hline 
Feb 25 & YAGO, DBpedia & Operation Abstention. \\ \hline
Feb 27 & YAGO, DBpedia, Wikidata$^\dagger$ & Action of 27 February 1941. \\ \hline
\end{tabular}

    \begin{tablenotes}
      \small
      \item $^\dagger$Wikidata misses the fact that this action is part of the World War II.
    \end{tablenotes}
  \end{threeparttable}
  
\end{table}

\textit{Assessing event popularity and relation strength in cross-cultural event-centric analytics:}
Event popularity and relation strength between events and entities vary across different cultural and linguistic contexts. 
For example, Table \ref{tab:top_3_events} presents the top-4 most popular events in the Russian vs. the English Wikipedia language editions as measured by how often these events are linked to in the respective Wikipedia edition. Whereas both Wikipedia language editions mention events of global importance, here the two World Wars, most frequently, other most popular events (e.g. ``October Revolution'' and ``American Civil War'') are language-specific.
The relation strength between events and entities in specific language contexts can be induced by counting their joint mentions in Wikipedia. 
For example, Table \ref{tab:ww_ii_relation_strength} lists the persons most related to the World War II in different language editions. 
Information regarding event popularity and relation strength can enable the selection of the most relevant timeline entries given the layout constraints (e.g. EventKG contains $2,816$ sub-events of the World War II).
An EventKG application to cross-lingual timeline generation is presented in \cite{gottschalk2018demo}. EventKG-empowered interfaces can be used as a starting point to identify controversial events for more detailed analysis using tools such as MultiWiki \cite{GottschalkD17}.

\begin{table}[t]
\centering
\footnotesize
    \caption{Most linked events in the English (en) and the Russian (ru) Wikipedia.}
    \label{tab:top_3_events}
\begin{tabular}{|l|l|r||l|r|}
\hline
\textbf{Rank} & \textbf{Event (en)} & \textbf{\#Links (en)} & \textbf{Event (ru)} & \textbf{\#Links (ru)} \\ \hline
1 & World War II & 189,716 & World War II & 25,295 \\ \hline
2 & World War I & 99,079 & World War I & 22,038 \\ \hline
3 & American Civil War & 37,672 & October Revolution & 7,533 \\ \hline
4 & FA Cup & 20,640 & Russian Civil War & 7,093 \\ \hline
\end{tabular}
\end{table}

 \begin{table}[t]
 \centering
 \footnotesize
 \caption{Top-3 persons mentioned jointly with the World War II per language.}
 \label{tab:ww_ii_relation_strength}
\begin{tabular}{l|l|l|l|l|}
 \cline{2-5}
  & \textbf{fr} & \textbf{de} & \textbf{ru} & \textbf{pt} \\ \hline
 \multicolumn{1}{|l|}{\textbf{1}} & Adolf Hitler & Adolf Hitler & Adolf Hitler & Adolf Hitler \\ \hline
 \multicolumn{1}{|l|}{\textbf{2}} & Charles de Gaulle & Winston Churchill & Franklin D. Roosevelt & Getúlio Vargas \\ \hline
 \multicolumn{1}{|l|}{\textbf{3}} & Winston Churchill & Franklin D. Roosevelt & Joseph Stalin & Joseph Stalin \\ \hline

 \end{tabular}
 \end{table}

\textit{Impact in supporting the adoption of Semantic Web technologies:}
EventKG follows best practices in data publishing and relies on open data and W3C standards to make the data reusable for a variety of real-world applications.
We believe that researchers using EventKG outside the Semantic Web community, e.g. in the fields of NLP and Digital Humanities, will profit from the adoption of the W3C standards such as RDF, SPARQL and re-use of established vocabularies, thus stimulating adoption of Semantic Web technologies, e.g. in the context of Information Extraction, media analytics and cross-cultural studies.

\section{EventKG Data Model}
\label{sec:model}

\textit{The goal of the EventKG data model} is to facilitate a light-weight integration and fusion of heterogeneous event representations and temporal relations extracted from the reference sources, and make this information available to real-world applications. 
The EventKG data model is driven by the following goals:

\begin{itemize}
\item Define the key properties of events through a canonical representation.
\item Represent temporal relations between events and entities
(including event-entity, entity-event and entity-entity relations). 
\item Include information quantifying and further describing these relations.
\item Represent relations between events (e.g. in the context of event series).
\item Support an efficient light-weight integration of event representations and temporal relations originating from heterogeneous sources.
\item Provide provenance for the information included in EventKG.
\end{itemize}

\begin{figure}
 \centering
  \includegraphics[width=\textwidth]{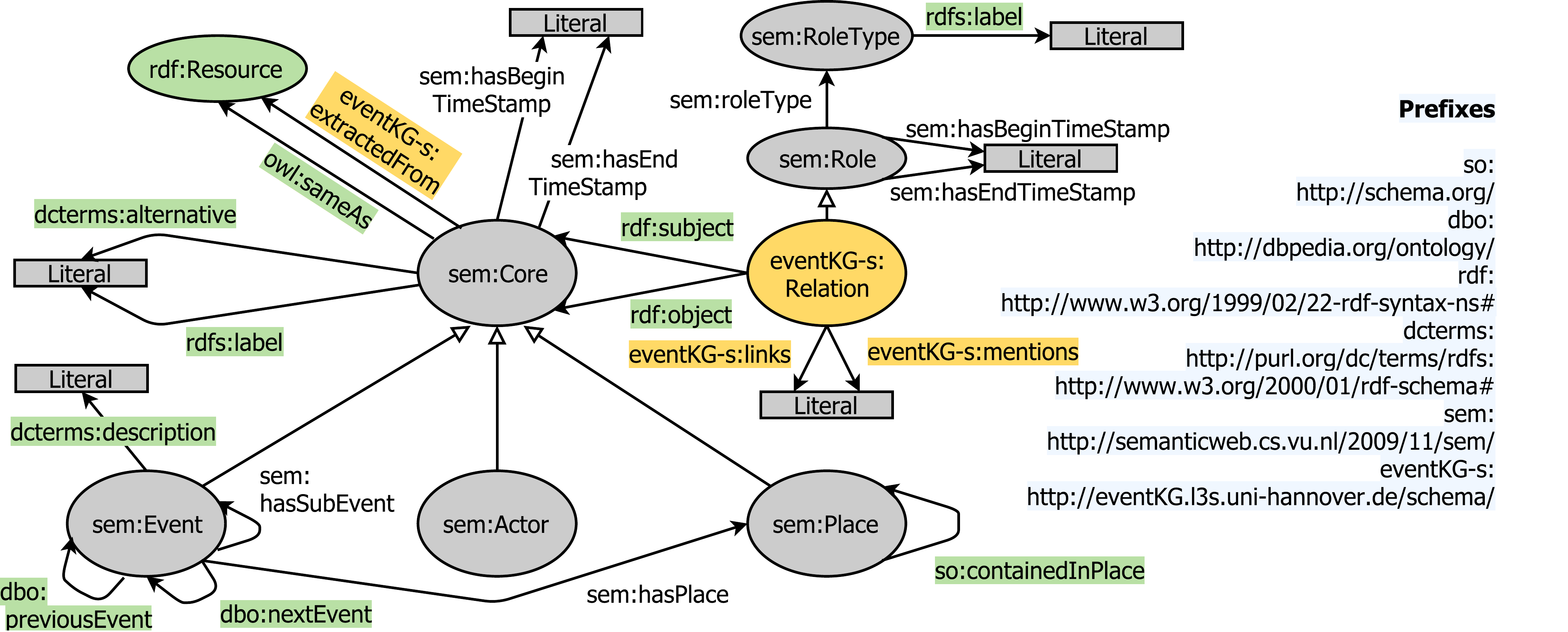}
  \caption{The EventKG schema based on SEM. Arrows with an open head denote \schema{rdfs:subClassOf} properties. Regular arrows visualize the \schema{rdfs:domain} and \schema{rdfs:range} restrictions on properties. Terms from other reused vocabularies are colored green. Classes and properties introduced in EventKG are colored orange.}
  \label{fig:schema}
\end{figure}

\textit{EventKG schema and the Simple Event Model:} In EventKG, we build upon the Simple Event Model (SEM) \cite{VanHage:2011} as a basis to model events. SEM is a flexible data model that provides a generic event-centric framework. 
Within the EvenKG schema (namespace \schema{eventKG-s}\footnote{\url{http://eventkg.l3s.uni-hannover.de/schema/}}), we adopt additional properties and classes to adequately represent the information extracted from the reference sources, to model temporal relations and event relations as well as to provide provenance information.
The schema of EventKG is presented in Fig. \ref{fig:schema}.

\textit{Events and entities}: 
SEM provides a generic event representation including  topical, geographical and temporal dimensions of an event, as well as links to its actors (i.e. entities participating in the event). Such resources are identified within the namespace \schema{eventKG-r}\footnote{\url{http://eventkg.l3s.uni-hannover.de/resource/}}.
Thus, the key classes of SEM and of the EventKG schema are \schema{sem:Event} representing events, \schema{sem:Place} representing locations and \schema{sem:Actor} to represent entities participating in events. Each of these classes is a subclass of \schema{sem:Core}, which is used to represent all entities in EventKG. (Note that entities in EventKG are not necessarily actors in the events; temporal relations between two entities are also possible). 
Events are connected to their locations through the \schema{sem:hasPlace} property. 
A \schema{sem:Core} instance can be assigned an existence time denoted via \schema{sem:has\-Begin\-Time\-Stamp} and \schema{sem:\-has\-End\-Time\-Stamp}. 
In addition to the SEM representation, EventKG provides textual information regarding events and entities extracted from the reference sources including labels (\schema{rdfs:label}), aliases (\schema{dcterms:\-alter\-native}) and descriptions of events (\schema{dcterms:\-description}). 

\begin{figure}
 \centering
  \includegraphics[width=0.85\textwidth]{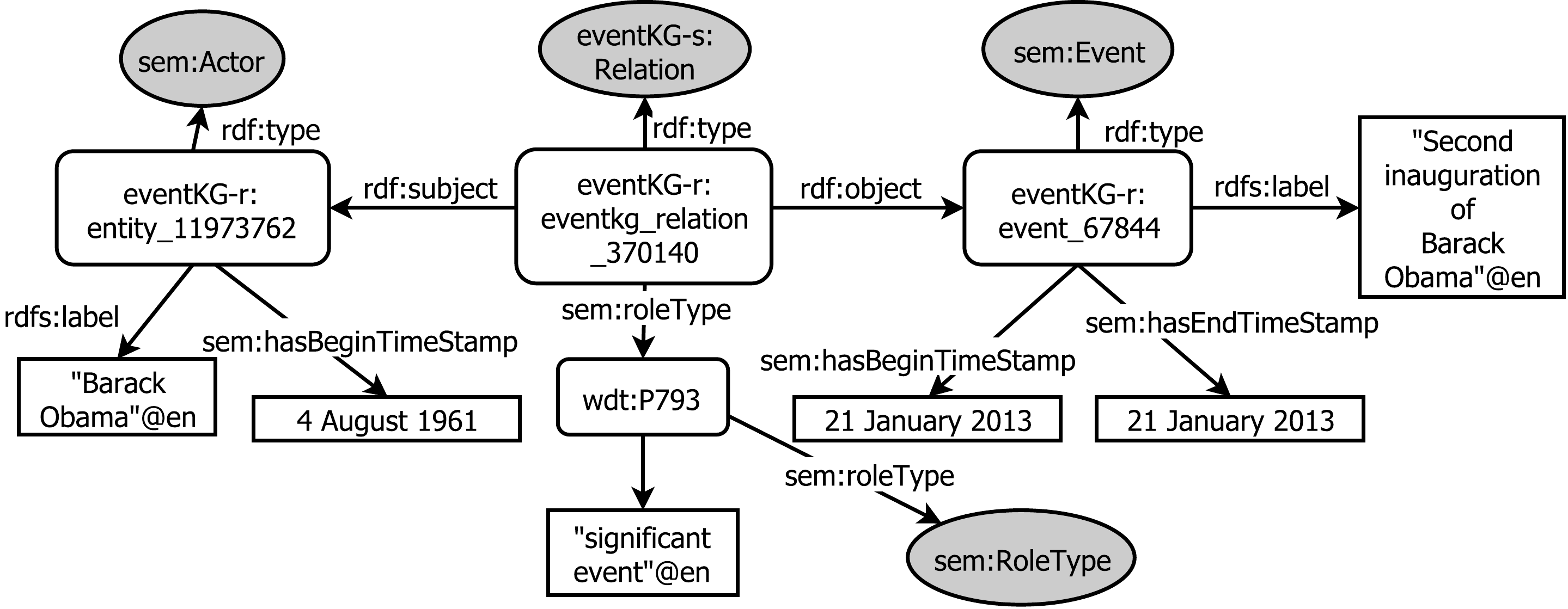}
  \caption{Example of the event representing the participation of Barack Obama in his second inauguration as a US president in 2013 as modeled in EventKG.}
  \label{fig:relation_instances}
\end{figure}

\textit{Temporal relations} are relations valid over a certain time period. In EventKG, they include event-entity, entity-event and entity-entity relations.
Temporal relations between events and entities typically connect an event and its actors (as in SEM).
A typical example of a temporal relation between two entities is a marriage. 
Temporal relations between entities can also indirectly capture information about events \cite{ROSPOCHER2016132}. For example, the DBpedia property \url{http://dbpedia.org/property/acquired} can be used to represent an event of acquisition of one company by another. 
Temporal relations in SEM are limited to the situation where an actor plays a specific role in the context of an event. This yields two limitations: (i) there is no possibility to model temporal relations between events and entities where the entity acts as a subject. For example, it is not possible to directly model the fact that ``Barack Obama'' participated in the event ``Second inauguration of Barack Obama'', as the entity ``Barack Obama'' plays the subject role in this relation; and (ii) a temporal relation between two entities such as a marriage can not be modeled directly. 
To overcome these limitations, EventKG introduces the class \schema{eventKG-s\-:\-Re\-lation} that links two \schema{sem:Core} instances (each representing an event or an entity). 
This relation can be annotated with a validity time and a property \schema{sem:\-Role\-Type} that characterizes the relation. This way, arbitrary temporal relations between entity pairs or relations involving an entity and an event can be represented. Fig. \ref{fig:relation_instances} visualizes the example mentioned above 
using the EventKG data model.

\textit{Relations with indirect temporal information}:
The temporal validity of a relation is not always explicitly provided, but can often be estimated based on the existence times of the participating entities or events.
For example, the validity of a ``mother'' relation can be determined using the birth date of the child entity.
Therefore, in addition to temporal relations with known validity times, EventKG also includes relations connected to events as well as relations connected to entities as long as the existence time of both entities is provided.

\textit{Other event and entity relations}: Relations between events (in particular sub-event, previous and next event relations)
play an important role in the context of event series (e.g. "Summer Olympics"), 
seasons containing a number of related events (e.g. in sports), or events related to a certain topic (e.g. operations in a military conflict).
Sub-event relations are modeled using the \schema{so:\-has\-Sub\-Event} property. To interlink events within an event series such as the sequence of Olympic Games, the properties \schema{dbo:\-previous\-Event} and \schema{dbo:\-next\-Event} are used. 
A location hierarchy is provided through the property \schema{so:con\-tained\-InPlace}.

\textit{Towards measuring relation strength and event popularity}:
Measuring relation strength between events and entities and event popularity enables answering question like \textit{``Who was the most important participant of the event $e$?''} or \textit{``What are the most popular events related to $e$?''}. 
We include two relevant factors in the EventKG schema: 

\textit{1. Links:} This factor represents how often the description of one entity refers to another entity. Intuitively, this factor can be used to estimate the popularity of the events and the strength of their relations.
In EventKG the links factor is represented through the predicate \schema{eventKG-s:\-links} in the domain of \schema{eventKG-s:\-Relation}.   
\schema{eventKG-s:\-links} denotes how often the Wikipedia article representing the relation subject links to the entity representing the  object.

\textit{2. Mentions:} \schema{eventKG-s:\-mentions} represents the number of relation mentions in external sources. Intuitively, this factor can be used to estimate the relation strength.
In EventKG, \schema{eventKG-s:\-mentions} denotes the number of sentences in Wikipedia that mention both, the subject and the object of the relation. 

\textit{Provenance information}: 
EventKG provides the following provenance information: 
(i) provenance of the individual resources; 
(ii) representation of the reference sources; and 
(iii) provenance of statements.

\textit{Provenance of the individual resources:} EventKG resources typically directly correspond to the events and entities contained in the reference sources (e.g. an entity representing Barack Obama in EventKG corresponds to the DBpedia resource \url{http://dbpedia.org/page/Barack_Obama}). In this case, the \schema{owl:sameAs} property is used to interlink both resources. 
EventKG resources can also be extracted from a resource collection. For example, philosophy events in 2007 can be extracted from the Wikipedia event list at \url{https://en.wikipedia.org/wiki/2007_in_philosophy}. 
In this case, the EventKG property \schema{eventKG-s:\-extracted\-From} is utilized to establish the link between the EventKG resource and the resource collection from which it was extracted.
Through the provenance URIs, background knowledge contained in the reference sources can be accessed.

\textit{Representation of the reference sources:} EventKG and each of the reference sources are represented through an instance of \schema{void:\-Dataset}\footnote{The VoID vocabulary \url{https://www.w3.org/TR/void/}.}. Such an instance in the namespace \schema{eventKG-g}\footnote{\url{http://eventkg.l3s.uni-hannover.de/graph/}} includes specific properties of the source (e.g. its creation date).

\textit{Provenance information of statements:}
A statement in EventKG is represented as a quadruple, containing a triple and a URI of the named graph it belongs to.
Through named graphs, EventKG offers an intuitive way to retrieve information extracted from the individual reference sources using SPARQL queries.

\section{EventKG Generation Pipeline}
\label{sec:extraction}
\begin{figure}[h]
  \centering
  \includegraphics[width=\textwidth]{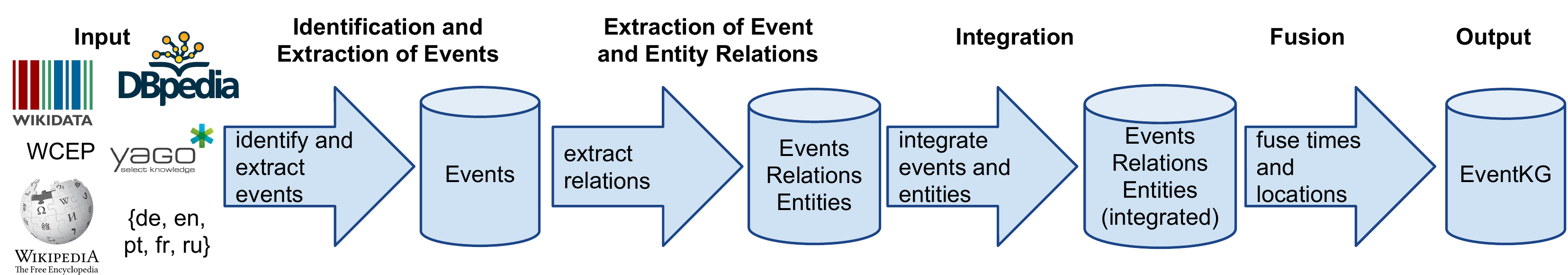}
  \caption{EventKG generation pipeline.}
  \label{fig:pipeline}
\end{figure}

The EventKG generation pipeline is shown in Fig. \ref{fig:pipeline}.

\textit{Input}: First, the dumps of the reference sources are collected.  

\textit{Identification and Extraction of Events}: Event instances are 
identified in the reference sources and extracted, as follows:

\textit{Step Ia: Identification and extraction of events.}

\begin{itemize}[leftmargin=*]
\item 
\textbf{Wikidata} \cite{Erxleben:2014}: We identify events as subclasses of Wikidata's ``event'' and ``occurrence''. The ``occurrence'' instances are added to increase recall. Some of the  identified subclasses are blacklisted manually. 
\item 
\textbf{DBpedia} \cite{dbpedia-swj}: For each language edition, we identify DBpedia events as instances of \schema{dbo:\-Event} or its subclasses.
\item 
\textbf{YAGO} \cite{Mahdisoltani:2014}: We do not use the YAGO ontology for event identification due to the noisy event subcategories (e.g. \schema{event} \textgreater\ \schema{act} \textgreater\ \schema{activity} \textgreater\ \schema{protection}\ \textgreater\ \schema{self-defense} \textgreater\ \schema{martial\_art}). YAGO events are identified in Step Ib.

\item 
\textbf{Wikipedia Event Lists}: For each language, we extract events from the Wikipedia event lists whose titles contain temporal expressions, such as ``2007 in Science'' and ``August 11'', using methods similar to \cite{Hienert:2012}.
\item 
\textbf{WCEP}: In the Wikipedia Current Events Portal, events are represented through rather brief textual descriptions and refer to daily happenings. We extract WCEP events using the WikiTimes interface \cite{Tran:2014}.
\end{itemize}

We manually evaluated a random sample of the events identified in this step in DBpedia and Wikidata including 100 events per KG and language edition, achieving precision of 98\% on average. 

\textit{Step Ib: Using additional event identification heuristics to increase recall.} 
First, we propagate the information regarding the identified events across the reference sources using existing \schema{owl:sameAs} links. 
Second, we use Wikipedia category names that match a manually defined language-dependent regular expression (e.g. English category names that end with `` events'') as an indication that a KG entry linked to such an article is an event. We manually evaluated this heuristic on a random sample of 100 events linked to the English and the Russian Wikipedia, respectively, achieving 94\% and 88\% precision. 

In EventKG V1.1, we do not explicitly distinguish between single events such as ``Solar eclipse of August 10, 1915'', seasons with a number of related events such as ``2008 Emperor's Cup'' and event series like ``Mario Marathon''. 

\textit{Extraction of Event and Entity Relations}:
We extract the following types of relations:
1) \textit{Temporal relations} are identified based on the availability of temporal validity information. 
Temporal relations are extracted from YAGO and Wikidata, as DBpedia does not provide such information.
2) \textit{Relations with indirect temporal information}:
we extract all relations involving events as well as relations of entities with known existence time.
3) \textit{Other event and entity relations}: we use a manually defined mapping table to identify predicates that represent event relations in EventKG such as 
\schema{so:\-has\-Sub\-Event} (e.g. we map Wikidata's \schema{part of} property (P361) to \schema{so:\-has\-Sub\-Event} in cases where the property is used to connect events), \schema{dbo:\-previous\-Event} and \schema{dbo:next\-Event} as well as \schema{so:\-con\-tained\-In\-Place} to extract location hierarchies.
We extract information that quantifies relation strength and event popularity based on the Wikipedia interlinking for each pair of interlinked entities containing at least one event. 
Entities are extracted only if they participate in an extracted relation. 

\textit{Integration}:
The statements extracted from the reference sources are included in the named graphs, each named graph corresponding to a reference source.
In addition, we create a named graph \schema{eventKG-g:event\_kg}.
Each \schema{sem:Event} and \schema{sem:Core} instance in the \schema{eventKG-g:event\_kg} integrates event-centric and entity-centric information from the reference sources related to equivalent real-world instances. 
For the instances extracted from the KGs, known \schema{owl:sameAs} links are used. Events extracted from the semi-structured sources are integrated using a rule-based approach based on descriptions, times and links. 

\textit{Fusion}: 
In the fusion step, we aggregate temporal, spatial and type information of \schema{eventKG-g:event\_kg} events using a rule-based approach. 
\textit{Location fusion}: For each event in \schema{eventKG-g:event\_kg}, we take the union of its locations from the different reference sources and exploit the \schema{so:\-con\-tained\-In\-Place} relations to reduce this set to the minimum (e.g. the set \{Paris, France, Lyon\} is reduced to \{Paris, Lyon\}).
\textit{Time fusion}: 
For each entity, event or relation with a known existence or a validity time stamp, the integration is done using the following rules: (i) ignore the dates at the beginning or end of a time unit (e.g. January, 1st), if alternative dates are available; (ii) apply a majority voting among the reference sources; (iii) take the time stamp from the trusted source (in order: Wikidata, DBpedia, Wikipedia, WCEP, YAGO).

\textit{Type fusion}: We provide
 \schema{rdf:type} information according to the DBpedia ontology (dbo), using types and \schema{owl:sameAs} links in the reference sources.

\textit{Output}: Finally, extracted instances and relations are represented in RDF according to the EventKG data model (see Section \ref{sec:model}). 
As described above, the information extracted from each reference source and the results of the fusion step are provided in separate named graphs.

\section{EventKG Characteristics}
\label{sec:characteristics}

In EventKG V1.1, we extracted event representations and relations in five languages 
from the latest available versions of each reference source as of 12/2017. 
Table \ref{tab:eventkg_overview} summarizes selected statistics from the EventKG V1.1, released in 03/2018. Overall, this version provides information for over $690$ thousand events and over $2.3$ million temporal relations.  
Nearly half of the events ($46.75\%$) originate from the existing KGs; the other half ($53.25\%$) is extracted from semi-structured sources. 
The data quality in the individual named graphs directly corresponds to the quality of the reference sources. 
In \schema{eventKG-g:event\_kg}, the majority of the events ($76.21\%$) possess a known start or end time. 
Locations are provided for $12.21\%$ of the events. The coverage of locations can be further increased in the future work, e.g. using NLP techniques to extract locations from the event descriptions.
Along with over $2.3$ million temporal relations, EventKG V1.1 includes relations between events and entities for which the time is not available. This results in overall over 88 million relations. 
Approximately half of these relations possess interlinking information.

\begin{table}[!t]
\centering
\footnotesize
\caption{Number of events and relations in \schema{eventKG-g:event\_kg}.}
\label{tab:eventkg_overview}
\begin{tabular}{l|r|r|r|}
\cline{2-4}
 & \multicolumn{1}{l|}{\textbf{\#Events}} & \multicolumn{1}{l|}{\textbf{Known time}} & \textbf{Known location} \\ \hline
\multicolumn{1}{|l|}{Events from KGs} & 322,669 & 163,977 & \multicolumn{1}{r|}{84,304} \\ \hline
\multicolumn{1}{|l|}{Events from semi-structured sources} & 367,578 & 362,064 & not extracted \\ \hline
\multicolumn{1}{|l|}{Relations} & 88,473,111 & 2,331,370 & not extracted\\ \hline
\end{tabular}
\end{table}

\subsection{Comparison of EventKG to its Reference Sources}
\label{sec:comparison}

We compare EventKG to its reference sources in terms of the number of the identified events and completeness of their representation.
The results of the event identification Step Ia are shown in Table \ref{tab:event_identification}. 
EventKG with $690,247$ events contains a significantly higher number of events than any of its reference sources. 
This is especially due to the integration of KGs and semi-structured sources.

\begin{table}[!t]
\scriptsize
\centering
\caption{Number of events extracted from the reference sources (Step Ia).}
\label{tab:event_identification}
\begin{tabular}{l|c|c|c|c|c|c|c|c|c|c|l}
\cline{2-11}
 & \multicolumn{5}{c|}{\textbf{DBpedia}} & \multicolumn{5}{c|}{\textbf{Wikipedia event lists}} &  \\ \cline{1-1} \cline{12-12} 
\multicolumn{1}{|c|}{\textbf{Wikidata}} & \textbf{en} & \textbf{fr} & \textbf{de} & \textbf{ru} & \textbf{pt} & \textbf{en} & \textbf{fr} & \textbf{de} & \textbf{ru} & \textbf{pt} & \multicolumn{1}{c|}{\textbf{WCEP}} \\ \hline
\multicolumn{1}{|r|}{266,198} & \multicolumn{1}{r|}{60,307} & \multicolumn{1}{r|}{43,495} & \multicolumn{1}{r|}{9,383} & \multicolumn{1}{r|}{5,730} & \multicolumn{1}{r|}{14,641} & \multicolumn{1}{r|}{131,774} & \multicolumn{1}{r|}{110,879} & \multicolumn{1}{r|}{21,191} & \multicolumn{1}{r|}{44,025} & \multicolumn{1}{r|}{18,792} & \multicolumn{1}{r|}{61,382} \\ \hline
\end{tabular}
\end{table}

Table \ref{tab:named_events_comparison} presents a comparison of the event representations in EventKG and 
its reference knowledge graphs (Wikidata, YAGO, DBpedia).
As we can observe, through the integration of event-centric information, EventKG: 
1) enables better event identification (e.g. we can map $322,669$ events from EventKG to Wikidata, 
whereas only $266,198$ were identified as events in Wikidata initially - see Table \ref{tab:event_identification}), and 
2) provides more complete event representations (i.e. EventKG provides a higher percentage of events with specified temporal and spatial information compared to Wikidata, that is the most complete reference sources). 
The most frequent event types are source-dependent (see Table \ref{tab:event_types}).

\begin{table}[!t]
\centering
\footnotesize
\caption{Comparison of the event representation completeness in the source-specific named graphs (after the Step Ib). 
}
\label{tab:named_events_comparison}
\begin{tabular}{lrrr|r|r|r|r|r|}
\cline{5-9}
 & \multicolumn{1}{l}{} & \multicolumn{1}{l}{} & \multicolumn{1}{l|}{} & \multicolumn{5}{c|}{\textbf{DBpedia}} \\ \cline{2-4}
\multicolumn{1}{l|}{} & \multicolumn{1}{c|}{\textbf{EventKG}} & \multicolumn{1}{c|}{\textbf{Wikidata}} & \multicolumn{1}{c|}{\textbf{YAGO}} & \multicolumn{1}{c|}{\textbf{en}} & \multicolumn{1}{c|}{\textbf{fr}} & \multicolumn{1}{c|}{\textbf{de}} & \multicolumn{1}{c|}{\textbf{ru}} & \multicolumn{1}{c|}{\textbf{pt}} \\ \hline
\multicolumn{1}{|l|}{\textbf{\#Events with}} & \multicolumn{1}{r|}{322,669} & \multicolumn{1}{r|}{322,669} & 222,325 & 214,556 & 78,527 & 62,971 & 47,304 & 35,682 \\ \hline
\multicolumn{1}{|l|}{\ Location (L)} & \multicolumn{1}{r|}{26.13\%} & \multicolumn{1}{r|}{11.70\%} & 26.61\% & 6.21\% & 8.32\% & 4.03\% & 10.60\% & 6.15\% \\ \hline
\multicolumn{1}{|l|}{\ Time (T)} & \multicolumn{1}{r|}{50.82\%} & \multicolumn{1}{r|}{33.00\%} & 39.02\% & 7.00\% & 17.21\% & 2.00\% & 1.35\% & 0.08\% \\ \hline
\multicolumn{1}{|l|}{\ L\&T} & \multicolumn{1}{r|}{21.97\%} & \multicolumn{1}{r|}{8.83\%} & 19.02 \% & 4.29\% & 0.00\% & 4.84\% & 1.18\% & 0.08\% \\ \hline
\end{tabular}
\end{table}

\begin{table}[!t]
 \centering
  \footnotesize
 \caption{The most frequent event types extracted from the references sources and the percentage of the events in that source with the respective type.
 }
 \label{tab:event_types}
 \begin{tabular}{lc|c|c|c|c|c|}
 \cline{3-7}
  & \multicolumn{1}{l|}{} & \multicolumn{5}{c|}{\textbf{DBpedia}} \\ \cline{2-2}
 \multicolumn{1}{l|}{} & \textbf{Wikidata} & \textbf{en} & \textbf{fr} & \textbf{de} & \textbf{ru} & \textbf{pt} \\ \hline
 \multicolumn{1}{|l|}{\textbf{dbo:type}} & season & \begin{tabular}[c]{@{}c@{}}Military\\ Conflict\end{tabular} & \begin{tabular}[c]{@{}c@{}}Sports\\ Event\end{tabular} & \begin{tabular}[c]{@{}c@{}}Tennis\\ Tournament\end{tabular} & \begin{tabular}[c]{@{}c@{}}Military\\ Conflict\end{tabular} & \begin{tabular}[c]{@{}c@{}}Soccer\\ Tournament\end{tabular} \\ \hline
 \multicolumn{1}{|l|}{\textbf{Events, \%}} & \multicolumn{1}{r|}{11.37\%} & \multicolumn{1}{r|}{6.31\%} & \multicolumn{1}{r|}{21.86\%} & \multicolumn{1}{r|}{33.00\%} & \multicolumn{1}{r|}{11.87\%} & \multicolumn{1}{r|}{16.17\%} \\ \hline
 \end{tabular}
 \end{table}

\subsection{Relation \& Fusion Statistics}
\label{sec:statistics}

Over 2.3 million temporal relations are an essential part of EventKG. 
The majority of the frequent predicates in EventKG such as ``member of sports team'' (882,398 relations), ``heritage designation'' (221,472), ``award received'' (128,125), and ``position held'' (105,333) originate from Wikidata. The biggest fraction of YAGO's temporal relations have the predicate ``plays for'' (492,263), referring to football players. Other YAGO predicates such as ``has won prize'' are less frequent. 
Overall, about $93.62\%$ of the temporal relations have a start time from 1900 to 2020.
$81.75\%$ of events extracted from KGs are covered by multiple sources.
At the fusion step, we observed that 93.79\% of the events that have a known start time agree on the start times across the different sources.

\subsection{Textual Descriptions}
\label{sec:lang_stat}

EventKG V1.1 contains information in five languages. 
Overall, $87.65\%$ of the events extracted from KGs provide an English label whereas only a small fraction ($4.49\%$) provide labels in all languages. 
Among the $367,578$ events extracted from the semi-structured sources, just $115$ provide a description in all five languages, e.g. the first launch of a Space Shuttle in 1981.
This indicates potential for further enrichment of  multilingual event descriptions in future work.

\section{Reusability Aspects}
\label{sec:reusability}

In order to facilitate an efficient reuse of EventKG, we provide the resource for download, as well as through a SPARQL endpoint. 
The homepage of EventKG provides a comprehensive documentation of the resource including example queries.  
A schema diagram of EventKG is presented in Fig. \ref{fig:schema}.
EventKG is modeled in RDF and is highly extensible. 
For example, it is possible to include further languages and customize the selection of the reference data sources.
Recent studies indicate that interlinking is an important factor of dataset reuse \cite{EndrisGTDZ0S17}.
To this extent, EventKG provides substantial interlinking with 
its reference sources. 

At the moment, the intended use of EventKG includes collaboration with EU-projects such as ALEXANDRIA (for enrichment of Web Archives with event-centric data)\footnote{\url{http://alexandria-project.eu/}}, and WDAqua ITN\footnote{WDAqua (Answering Questions using Web Data) \url{http://wdaqua.eu/}}, in the context of innovative event-centric Question Answering applications.
We believe that due to its unique nature and general applicability, EventKG will be widely reused by third parties in a number of communities in the future, as also discussed in Section \ref{sec:relevance}. 

EventKG follows best practices in data publishing. It uses 
RDF W3C standard to model and interlink the data it contains. 
EventKG adopts the open data and open source approach to make it available to a wide audience and facilitates data and software reuse. 
EventKG supports multilinguality of the data, provides dereferencable URIs and implements a persistent strategy to maintain its URIs across the versions, ensuring that the same URIs are consistently reused for the same real-world objects. 

EventKG reuses and extends an established event model, which is SEM \cite{VanHage:2011} to describe event-related information it includes,
and reuses existing vocabularies (e.g. DBpedia ontology, Dublin Core).
The EventKG metadata is provided using the VoID\footnote{\url{https://www.w3.org/TR/void/}} vocabulary.
EventKG follows FAIR principles\footnote{\url{https://www.force11.org/group/fairgroup/fairprinciples}} to make it findable, accessible, interoperable and reusable.
The EventKG description is available in human and machine readable formats at the EventKG homepage.

\section{Availability \& Sustainability}
\label{sec:availability_and_sustainability}

\textit{Availability Aspects}:
EventKG uses open standards and is publicly available under a persistent URI\footnote{\url{https://doi.org/10.5281/zenodo.1112283}} under the CC BY 4.0 license\footnote{\url{https://creativecommons.org/licenses/by/4.0/}}.
The EventKG homepage\footnote{\url{http://eventKG.l3s.uni-hannover.de}} provides information on citing the resource. 
Our extraction pipeline is available as open source software on GitHub\footnote{\url{https://github.com/sgottsch/eventkg}} under the MIT License\footnote{\url{https://opensource.org/licenses/MIT}}.

\textit{Sustainability Plan}:
The sustainability of EventKG is ensured though three building blocks:
1) \textit{Open source architecture and software}: The software developed for the creation of EventKG is available as open source and can be re-used by the community to extract a new version of the knowledge graph, or extend the resource to include more reference sources, languages, or event properties.
2) \textit{Integration of existing publicly available data}: The reference sources that serve as a basis for the data within the EventKG are publicly available and many of them are maintained by the community, so that it is possible to maintain a fresh version of the resource, in particular to include new events. 
3) \textit{Maintenance of EventKG}: The authors plan to 
perform regular EventKG updates.
The URIs of EventKG resources will be maintained and remain stable across versions.

\section{Related Work}
\label{sec:related}

\textit{Data models and vocabularies for events:} Several data models and the corresponding vocabularies (e.g. \cite{Shaw:2013,ROSPOCHER2016132,VanHage:2011,Guha:2011}) provide means to model events. For example, the ECKG model proposed by Rospocher et al. \cite{ROSPOCHER2016132} enables fine-grained textual annotations to model events extracted from news collections. 
The Simple Event Model (SEM) \cite{VanHage:2011}, schema.org \cite{Guha:2011} and the Linking Open Descriptions of Events (LODE) ontology \cite{Shaw:2013} provide means to describe events and interlink them with actors, times and places. 
In EventKG, we build upon SEM and extend this model to represent a wider range of temporal relations and to provide additional information regarding events. 

\textit{Extracting event-centric information:}
Most approaches for automatic knowledge graph construction and integration focus on entities and related facts rather than events. Examples include DBpedia \cite{dbpedia-swj}, Freebase \cite{Bollacker:2008}, 
YAGO \cite{Mahdisoltani:2014} and YAGO+F \cite{Demidova:2013}. 
In contrast, EventKG is focused on events and temporal relations. 
In \cite{Tran:2014}, the authors extract event information from WCEP. 
EventKG builds upon this work to include WCEP events. 

\textit{Extraction of events and facts from news:} Recently, the problem of building knowledge graphs directly from plain text news \cite{ROSPOCHER2016132}, and extraction of named events from news \cite{Kuzey:2014} have been addressed. These approaches apply Open Information Extraction methods and develop them further to address specific challenges in the event extraction in the news domain. State-of-the-art works that automatically extract events from news potentially obtain noisy and unreliable results (e.g. the state-of-the-art extraction approach in \cite{ROSPOCHER2016132} reports an accuracy of only 0.551). In contrast, contemporary events included in EventKG originate from manually curated sources such as WCEP and Wikipedia event lists.

\section{Conclusion}
\label{sec:conclusion}

In this paper we presented EventKG -- a multilingual knowledge graph that integrates and harmonizes event-centric and temporal information regarding historical and contemporary events. EventKG V1.1 includes over 690 thousand event resources and over 2.3 million temporal relations.
Unique features of EventKG include light-weight integration and fusion of structured and semi-structured multilingual event representations and temporal relations in a single knowledge graph, as well as the 
provision of information to facilitate 
assessment of relation strength and event popularity, while providing provenance.
The light-weight integration enables to significantly increase the coverage and completeness of the included event representations, in particular with respect to times and locations.

\bibliographystyle{splncs04}


{\footnotesize

\subsubsection*{Acknowledgements} This work was partially funded by  the ERC ("ALEXANDRIA", 339233) and BMBF ("Data4UrbanMobility", 02K15A040).

\bibliography{references}

}

\end{document}